\documentclass[conference]{IEEEtran}
\IEEEoverridecommandlockouts
\usepackage{cite}
\usepackage{amsmath,amssymb,amsfonts}
\usepackage{algorithmic}
\usepackage{graphicx}
\usepackage{textcomp}
\usepackage{xcolor}
\usepackage{booktabs}
\usepackage{multirow}
\usepackage{graphicx}
\usepackage{hyperref}   
\usepackage{xurl}
\hypersetup{
    colorlinks=true,
    linkcolor=black,        
    citecolor=black,        
    filecolor=black,        
    urlcolor=black,       
    breaklinks=true
}
\usepackage{balance}

\makeatletter
\newcommand{\linebreakand}{%
  \end{@IEEEauthorhalign}
  \hfill\mbox{}\par
  \mbox{}\hfill
  \begin{@IEEEauthorhalign}
}
\makeatother

\definecolor{improve}{RGB}{0,128,0}    
\definecolor{degrade}{RGB}{255,0,0}    

\definecolor{improve}{RGB}{0,128,0}    
\definecolor{degrade}{RGB}{255,0,0}    

\def\BibTeX{{\rm B\kern-.05em{\sc i\kern-.025em b}\kern-.08em
    T\kern-.1667em\lower.7ex\hbox{E}\kern-.125emX}}
\begin{document}

\title{\huge Hate Speech Detection using Large Language Models with Data Augmentation and Feature Enhancement}

\author{\IEEEauthorblockN{1\textsuperscript{st} Brian Jing Hong Nge}
\IEEEauthorblockA{\textit{AiLECS Lab} \\
\textit{Monash University}\\
Melbourne, Australia \\
bnge0001@student.monash.edu}

\and
\IEEEauthorblockN{2\textsuperscript{nd} Stefan Su}
\IEEEauthorblockA{\textit{AiLECS Lab} \\
\textit{Monash University}\\
Melbourne, Australia \\
ssuu0017@student.monash.edu}

\and
\IEEEauthorblockN{3\textsuperscript{rd} Thanh Thi Nguyen}
\IEEEauthorblockA{\textit{AiLECS Lab} \\
\textit{Monash University}\\
Melbourne, Australia \\
thanh.nguyen9@monash.edu\textsuperscript{*}\thanks{\textsuperscript{*}Corresponding author.}}

\linebreakand
\IEEEauthorblockN{4\textsuperscript{th} Campbell Wilson}
\IEEEauthorblockA{\textit{AiLECS Lab} \\
\textit{Monash University}\\
Melbourne, Australia \\
campbell.wilson@monash.edu}

\and
\IEEEauthorblockN{5\textsuperscript{th} Alexandra Phelan}
\IEEEauthorblockA{\textit{School of Social Sciences} \\
\textit{Monash University}\\
Melbourne, Australia \\
Alexandra.R.Phelan@monash.edu}

\and
\IEEEauthorblockN{6\textsuperscript{th} Naomi Pfitzner}
\IEEEauthorblockA{\textit{School of Social Sciences} \\
\textit{Monash University}\\
Melbourne, Australia \\
Naomi.Pfitzner@monash.edu}
}

\maketitle

\begin{abstract}
This paper evaluates data augmentation and feature enhancement techniques for hate speech detection, comparing traditional classifiers, e.g., Delta Term Frequency-Inverse Document Frequency (Delta TF-IDF), with transformer-based models (DistilBERT, RoBERTa, DeBERTa, Gemma-7B, gpt-oss-20b) across diverse datasets. It examines the impact of Synthetic Minority Over-sampling Technique (SMOTE), weighted loss determined by inverse class proportions, Part-of-Speech (POS) tagging, and text data augmentation on model performance. The open-source gpt-oss-20b consistently achieves the highest results. On the other hand, Delta TF-IDF responds strongly to data augmentation, reaching 98.2\% accuracy on the Stormfront dataset. The study confirms that implicit hate speech is more difficult to detect than explicit hateful content and that enhancement effectiveness depends on dataset, model, and technique interaction. Our research informs the development of hate speech detection by highlighting how dataset properties, model architectures, and enhancement strategies interact, supporting more accurate and context-aware automated detection.
\end{abstract}

\begin{IEEEkeywords}
hate speech detection, data augmentation, feature enhancement, transformer models, BERT, gpt-oss-20b, class imbalance
\end{IEEEkeywords}

\section{Introduction}
The rise of online platforms has enabled extremist groups to disseminate harmful content at scale. In Australia, far-right terrorism investigations have increased drastically, with the Australia Security Intelligence Organisation (ASIO) confirming that in 2021, about 50\% of its domestic caseload was looking at far-right groups, from about 10\% before 2019 \cite{zwartz2020australian}. Online environments have an impact on the increase of such extremist ideologies, with the Australian Government recently classifying the online network Terrorgram, as a terrorist organisation \cite{australia2025terrorgram}.
Traditional monitoring methods often rely on manual review and filters. Although automated hate speech detection and moderation systems deployed by major platforms such as Meta’s Facebook and Instagram, YouTube and TikTok have been able to remove large quantities of explicit hateful content, evidenced by the fact that these platforms are not cesspools of hate, and have committed to improved review processes under regulatory frameworks like the EU’s Digital Services Act, there exists gaps for improvement. These gaps arise from perpetrators using more sophisticated and implicit communication tactics such as, subtler and context-dependent speech, as well as the sheer volume of data involved~\cite{conway2006terrorism}. The studies in \cite{mullah2021advances,badjatiya2017deep} demonstrated the viability of using machine learning to detect hate speech, which is supported by open-source models such as BERT and GPT offering strong baselines for semantic classification. 

Recent reviews in \cite{albladi2025hate,aljawazeri2024addressing} detail advancements in using Large Language Models (LLMs) such as better contextual understanding, better performance in generalisation due to pre-training, and ease of adaptability to downstream tasks and other languages. Additionally, they also provide challenges for the use of LLMs such as availability of quality datasets and possible biases from the pre-training data. 

Existing research highlights key gaps limiting real-world hate speech detection. First, access to hate speech labelled is often restricted, with only a limited number of high-quality openly available datasets. Sources of openly available datasets include Stormfront and pro-ISIS Twitter \cite{de2018hate}, Hate Corpus \cite{elsherief2021latent} and the Gab \& Reddit Dataset \cite{qian2019benchmark}. Conversely, an example of a restricted dataset is the U.S. Extremist Crime Database (ECBD) \cite{freilich2014introducing}, with further datasets being restricted due to their data having to conform to platform specific terms and conditions. These resources highlight the importance of diverse, well-annotated and accessible datasets for advancing research in this field.

Second, the ratio of hate-speech to non-hate-speech examples remains a persistent challenge, with hate speech typically representing a small proportion of overall content, which we refer to as class imbalance, resulting in poor recall for minority hate speech class \cite{fernandez2018imbalanced}. 

Third, while transformer models are performant, there is still a lack of comprehensive evaluations spanning the full spectrum of model architectures, from traditional classifiers to advanced LLMs. 
Most studies focus on narrow model families, making it difficult to draw general conclusions about relative effectiveness across traditional methods and advanced LLMs. 

Fourth, recent data augmentation advances, as documented by Bayer et al. \cite{bayer2022survey}, show promise for addressing dataset limitations and improving model robustness, hence there is a gap in research to utilise these new techniques.

This paper addresses these limitations through an evaluation of traditional classifiers, e.g., Delta Term Frequency-Inverse Document Frequency (Delta TF-IDF), established transformer architectures (DistilBERT, RoBERTa, DeBERTa), and modern LLMs (Gemma-7B, gpt-oss-20b) for hate speech detection. We investigate class imbalance mitigation strategies including Synthetic Minority Over-sampling Technique (SMOTE), class weighting, and data augmentation approaches. Our evaluation includes advanced preprocessing methods including Part-of-Speech (POS) tagging and multiple data augmentation strategies to improve model generalization.
This paper addresses a critical research question: How do data augmentation and feature enhancement techniques affect hate speech detection performance across traditional classifiers and LLMs when applied to datasets with varying levels of implicit and explicit hateful content? 
The primary contribution of this work lies in a comprehensive empirical benchmarking and the implementation code is available at: https://github.com/Brian3410/HateSpeechDetection.

\section{Related Work}
\subsection{Hate Speech Detection Approaches}
Hate speech detection has evolved from basic rule-based keyword matching to advanced machine learning methods. Initial research used lexical characteristics and bag-of-words models, where Support Vector Machines and Naive Bayes algorithms demonstrated early potential \cite{pen2024comparative}. The integration of TF-IDF weighting mechanisms improved feature representation by prioritizing discriminative terms while diminishing the influence of commonly occurring words.
Modern developments in deep learning have fundamentally transformed hate speech detection capabilities. Convolutional Neural Networks (CNNs) exhibited proficiency in identifying local textual patterns, while Recurrent Neural Networks (RNNs) and Long Short-Term Memory (LSTM) architectures proved more effective for capturing sequential text dependencies. However, recent studies indicate deep learning is not always superior, with traditional machine learning techniques still demonstrating competitive performance in certain contexts \cite{pen2024comparative}. The emergence of transformer-based models, such as BERT and LLMs \cite{nguyen2024fine}, marked a breakthrough by enabling bidirectional context understanding and transfer learning from large pretrained datasets.

\subsection{Class Imbalance in Hate Speech Detection}
Class imbalance is a major challenge in hate speech detection, as non-hateful content greatly outnumbers hateful instances. Traditional approaches include oversampling methods like SMOTE, undersampling, and cost-sensitive learning that assign different penalties to classes. Addressing class imbalance has become increasingly important in machine learning, particularly in domains like social media analysis~\cite{fernandez2018smote}.
Recent research has explored advanced resampling strategies specifically designed for text classification. The survey in~\cite{bayer2022survey} provides a comprehensive coverage of text augmentation methods, categorizing approaches into character-level, word-level, and sentence-level transformations.

\subsection{Feature Enhancement Techniques}
POS tagging has emerged as a valuable preprocessing step for hate speech detection, providing grammatical context that improves model understanding of linguistic patterns. However, POS tagging can suffer from position bias, where model performance drops if tokens appear in unfamiliar positions during training \cite{ben2024impact}. 
Data augmentation techniques for text classification have gained significant attention following the success of Easy Data Augmentation (EDA) methods \cite{pellicer2023data}. These include synonym replacement, random insertion, swapping, and deletion to increase dataset diversity while preserving meaning. More sophisticated augmentation methods leverage language models to generate contextually appropriate variations of training samples, with recent advances including contextual augmentation using BERT-based models and paraphrasing techniques using T5 and BART architectures.

\section{Methodology}
\subsection{Datasets}
Our evaluation covers four hate speech datasets with varied linguistic patterns and annotations.
\begin{itemize}
\item \textbf{Hate Corpus Dataset:} 
    \begin{itemize}
        \item A dataset made to fulfill the need for an implicit hate speech and benchmark corpus \cite{elsherief2021latent}.
        \item Comprehensive dataset comprising 34,183 tweets from prominent extremist groups in the United States, with 8,188 tweets labeled as implicit hate speech \cite{elsherief2021latent}.
    \end{itemize}

\item \textbf{Gab \& Reddit Dataset:}
    \begin{itemize}
        \item A dataset comprised of examples obtained from Gab, a known far-right social network, and Reddit, a popular discussion platform \cite{qian2019benchmark}.
        \item Conversational dataset from alternative social media platforms, comprising 35,249 no hate and 19,862 hate instances \cite{qian2019benchmark}.
    \end{itemize}
    
\item \textbf{Stormfront Dataset:} 
    \begin{itemize}
        \item A dataset comprised of examples obtained from Stormfront, a known white supremacist online forum.
        \item Sentence-level annotated dataset containing 10,944 sentences from a white supremacist forum, with 1,196 containing explicit hate speech \cite{de2018hate}.
    \end{itemize}
    
\item \textbf{Merged Dataset:}
    \begin{itemize}
        \item A unified dataset combining over 54,680 samples from the aforementioned datasets.
    \end{itemize}
\end{itemize}

\subsection{Model Architectures}
\begin{itemize}
    \item \textbf{Delta TF-IDF:} The delta variant improves the traditional bag-of-words approach with TF-IDF weighting by incorporating feature selection based on discriminative power for hate speech classification \cite{martineau2009improving}.
    \item \textbf{DistilBERT:} Distilled version of BERT that is smaller and faster. It maintains 97\% of BERT's language understanding capabilities while reducing model size by 40\% and achieving 60\% faster inference times \cite{sanh2020distilbert}.
    \item \textbf{RoBERTa:} Robustly optimized BERT pretraining approach that improves upon BERT by removing the Next Sentence Prediction task, using dynamic masking, larger batches, and longer training on diverse data \cite{liu2019roberta,patel2025misinformation}.
    \item \textbf{Gemma-7B:} Google's latest instruction-tuned language model designed for safe, responsible AI applications, offering state-of-the-art performance on various NLP tasks including text classification \cite{team2024gemma,nguyen2025large}.
    \item \textbf{DeBERTaV3:} The third iteration of DeBERTa, which was based on RoBERTa, with the incorporation of disentangled attention and the increase of parameters to 1.5 billion \cite{he2021debertav3}.
    \item \textbf{gpt-oss-20b:} OpenAI’s open-source reasoning model with 20 billion parameters with Chain-of-Thought (CoT) capabilities \cite{agarwal2025gptoss}.
\end{itemize}

\section{Techniques and Training Configurations}
\begin{figure*}[htbp]
\centerline{\includegraphics[width=2\columnwidth]{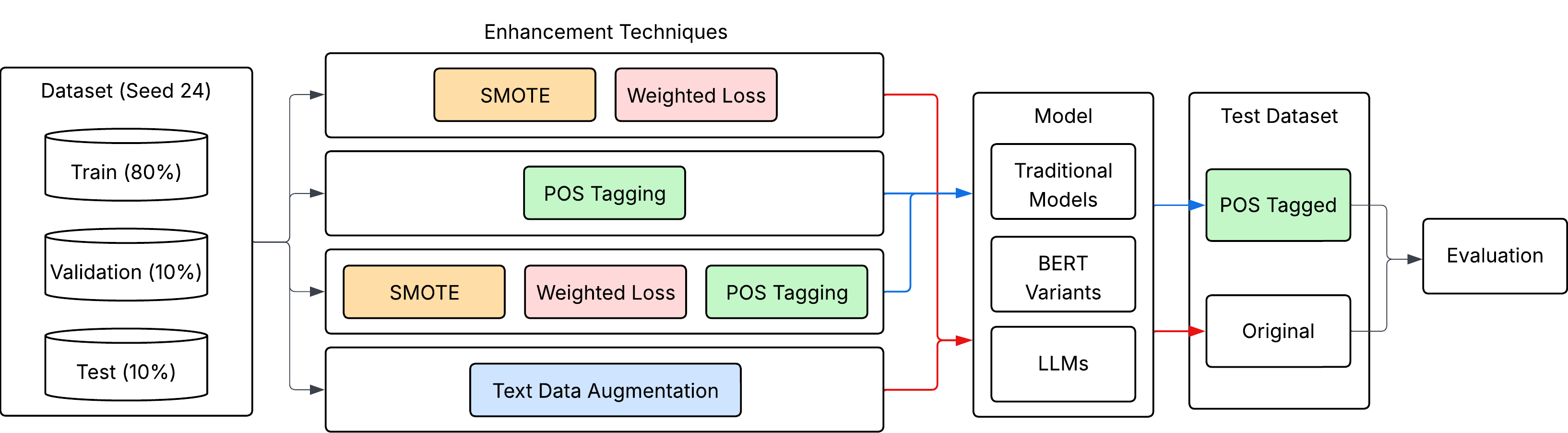}}
\caption{Training \& evaluation process for each dataset.}
\label{fig1}
\end{figure*}

We evaluate model performance across all approaches (Delta TF-IDF, DistilBERT, RoBERTa, Gemma-7B, DeBERTaV3, gpt-oss-20b) under four distinct enhancement technique configurations as presented in Fig.~\ref{fig1}.

\subsection{Baseline Configuration}
\textbf{Standard Training:} Establishes performance benchmarks using each model's default architecture without additional enhancement techniques. This configuration employs standard training procedures over 10 epochs with stratified data splits (80\% training, 10\% validation, 10\% testing) using random seed 42 to ensure reproducible results across all experimental runs and model architectures. For neural models, training utilizes conventional optimization strategies including standard cross-entropy loss, balanced sampling approaches, and model-specific default hyperparameters. Traditional models use their standard feature extraction and classification procedures. For Delta TF-IDF, standard training uses 10 fold stratified cross-validation on the 80\% training data. This approach simulates the 10 epoch training of neural models by conducting 10 independent training runs on different data partitions, maintaining class balance and providing comparable evaluation rigor and statistical robustness. This methodology ensures a fair, consistent performance comparison across model architectures.

\subsection{Linguistic Enhancement Configuration}
\label{sec:linguistic}
\textbf{SMOTE \& Weighted Loss:} Application of SMOTE to training data combined with class-weighted loss functions to address imbalance during model training \cite{fernandez2018smote}. The configuration applies SMOTE resampling using intelligent TF-IDF vectorization (1000 features, unigrams+bigrams) with cosine similarity-based synthetic sample mapping to training data exclusively. Dynamically computed class weights are used to adjust loss function penalties to account for class frequency disparities. For neural architectures, a specialized WeightedTrainer handles weighted cross-entropy loss computation, while traditional models integrate balanced sampling strategies directly into their training procedures.

\textbf{POS Feature Integration:} Incorporates grammatical structure awareness through systematic extraction of linguistic ratio features (noun, verb, adjective, adverb, pronoun, interjection ratios, plus profanity and capitalization patterns) using spaCy's English language model \cite{ben2024impact}. For neural architectures, POS features undergo dimensionality projection and concatenation with model representations through additional classifier layers, while traditional models integrate these features directly into their feature vectors. This technique enables models to leverage grammatical patterns indicative of hate speech discourse, addressing potential position bias issues that can affect token classification performance when linguistic features appear in under-represented positions during training \cite{ben2024impact}.

\subsection{Comprehensive Balancing Configuration}
\textbf{SMOTE Oversampling \& Weighted Loss \& POS Integration:} The combination of the aforementioned techniques (presented in Section~\ref{sec:linguistic}) addressing both class imbalance and linguistic feature representation. 

\subsection{Data Diversity Enhancement Configuration}
\textbf{Text Data Augmentation:} Implements sophisticated multi-layered text transformation techniques exclusively during training to enhance dataset diversity and model robustness~\cite{pellicer2023data}. The augmentation framework combines word-level transformations (WordNet synonym replacement, random swapping/insertion/deletion with hate-speech-aware preservation), character-level modifications (keyboard neighbor replacement, spelling error simulation, realistic typo introduction), sentence-level restructuring (contraction transformation, structural variations), and contextual augmentation (transformer-based paraphrasing with semantic similarity validation). Quality control keeps important hate speech words and group names unchanged while creating more examples of the minority class through text variations to balance the dataset.

\section{Evaluation Metrics}
Model performance is assessed using diverse evaluation metrics designed to capture different classification aspects. This is particularly important for imbalanced datasets where traditional accuracy alone may be misleading. All metrics are computed using scikit-learn's standard implementations, with results reported as mean values to ensure statistical reliability and reproducibility of findings.

\textbf{Accuracy:} Measures the overall proportion of correct predictions across all classes, providing a general performance baseline. This metric can be misleading in imbalanced datasets.
\begin{equation}
\text{Accuracy} = \frac{\text{TP} + \text{TN}}{\text{TP} + \text{TN} + \text{FP} + \text{FN}}
\end{equation}

\textbf{Macro F1:} Calculates the unweighted average of F1 scores across all classes. This balances performance evaluation between classes. Unlike micro F1, it prevents class imbalance from obscuring minority class performance.
\begin{equation}
F_1 = \frac{2 \times \text{Precision} \times \text{Recall}}{\text{Precision} + \text{Recall}}
\end{equation}

\textbf{F$_{0.5}$ Score:} A variant of the F1 score that weights precision more heavily than recall ($\beta=0.5$), making it suitable for applications where minimizing false positives is prioritized. In hate speech detection, this metric is valuable when the cost of incorrectly flagging non-hateful content as hateful is high.
\begin{equation}
F_\beta = \frac{(1 + \beta^2) \times \text{Precision} \times \text{Recall}}{\beta^2 \times \text{Precision} + \text{Recall}}
\end{equation}

\textbf{F$_2$ Score:} Conversely weights recall more heavily than precision ($\beta=2$), emphasizing the model's ability to identify all instances of hate speech. This metric is critical when missed hate speech poses risks to user safety and moderation.

\textbf{AUC (Area Under the ROC Curve):} Evaluates the model's ability to distinguish between classes across all thresholds, offering insights into the model's discriminative capability independent of specific decision boundaries.

\textbf{Weighted F1:} Calculates the F1 score for each class and then computes their average weighted by the number of true instances for each class. This approach accounts for class imbalance by giving more importance to classes with larger support, providing a performance measure that reflects the dataset's natural distribution.

\textbf{Weighted Precision:} Measures the weighted average of precision scores across classes, emphasizing performance on more frequent classes while still considering minority class performance.

\textbf{Weighted Recall:} Computes the weighted average of recall scores, indicating the model's capability to identify true instances across all classes while accounting for class frequency.

\section{Results and Discussions}
Our evaluation reveals significant variation in model performance across datasets and enhancement techniques. The transformer-based models (DistilBERT, RoBERTa, DeBERTa, Gemma-7B, gpt-oss-20b) consistently outperformed the traditional Delta TF-IDF approach across all datasets, with gpt-oss-20b achieving the highest overall performance.

\begin{table*}[htbp]
\centering
\caption{SMOTE \& Weighted Loss Configuration Results}
\label{tab:smote_weighted}
\resizebox{\textwidth}{!}{
\begin{tabular}{llcccccc}
\toprule
\textbf{Dataset} & \textbf{Metric} & \textbf{Delta TF-IDF} & \textbf{DistilBERT} & \textbf{DeBERTaV3} & \textbf{RoBERTa} & \textbf{gpt-oss-20b} & \textbf{Gemma-7B} \\
\midrule
\multirow{3}{*}{Hate Corpus} 
& Accuracy & 0.655 (\textcolor{degrade}{0.566}) & 0.694 (\textcolor{degrade}{0.475}) & 0.712 (\textcolor{degrade}{0.507}) & 0.738 (\textcolor{degrade}{0.488}) & 0.757 (\textcolor{degrade}{0.695}) & 0.728 (\textcolor{degrade}{0.596}) \\
& Macro F1 & 0.412 (\textcolor{degrade}{0.382}) & 0.444 (\textcolor{degrade}{0.369}) & 0.487 (\textcolor{degrade}{0.418}) & 0.480 (\textcolor{degrade}{0.396}) & 0.483 (\textcolor{improve}{0.566}) & 0.490 (\textcolor{improve}{0.540}) \\
& AUC & 0.312 (\textcolor{improve}{0.345}) & 0.400 (\textcolor{degrade}{0.336}) & 0.524 (\textcolor{degrade}{0.456}) & 0.561 (\textcolor{degrade}{0.401}) & 0.518 (\textcolor{improve}{0.564}) & 0.592 (\textcolor{improve}{0.597}) \\
\midrule
\multirow{3}{*}{Gab Reddit} 
& Accuracy & 0.889 (\textcolor{improve}{0.890}) & 0.908 (\textcolor{degrade}{0.907}) & 0.909 (\textcolor{degrade}{0.907}) & 0.910 (\textcolor{degrade}{0.909}) & 0.919 (\textcolor{degrade}{0.916}) & 0.856 (\textcolor{improve}{0.859}) \\
& Macro F1 & 0.874 (\textcolor{improve}{0.876}) & 0.900 (\textcolor{degrade}{0.899}) & 0.903 (\textcolor{degrade}{0.901}) & 0.902 (0.902) & 0.912 (\textcolor{degrade}{0.909}) & 0.842 (\textcolor{improve}{0.849}) \\
& AUC & 0.918 (\textcolor{degrade}{0.914}) & 0.925 (\textcolor{improve}{0.943}) & 0.952 (\textcolor{improve}{0.954}) & 0.949 (\textcolor{degrade}{0.945}) & 0.911 (0.911) & 0.923 (\textcolor{improve}{0.924}) \\
\midrule
\multirow{3}{*}{Stormfront} 
& Accuracy & 0.897 (0.897) & 0.921 (\textcolor{improve}{0.922}) & 0.922 (\textcolor{improve}{0.932}) & 0.931 (\textcolor{degrade}{0.919}) & 0.932 (\textcolor{degrade}{0.911}) & 0.911 (\textcolor{degrade}{0.863}) \\
& Macro F1 & 0.558 (0.558) & 0.772 (\textcolor{degrade}{0.766}) & 0.802 (\textcolor{improve}{0.821}) & 0.811 (\textcolor{degrade}{0.776}) & 0.815 (\textcolor{degrade}{0.785}) & 0.733 (\textcolor{improve}{0.735}) \\
& AUC & 0.819 (0.819) & 0.839 (\textcolor{improve}{0.887}) & 0.924 (\textcolor{degrade}{0.909}) & 0.846 (\textcolor{improve}{0.932}) & 0.794 (\textcolor{improve}{0.800}) & 0.911 (\textcolor{improve}{0.921}) \\
\midrule
\multirow{3}{*}{Merged Dataset} 
& Accuracy & 0.821 (\textcolor{improve}{0.822}) & 0.855 (\textcolor{improve}{0.860}) & 0.872 (\textcolor{degrade}{0.867}) & 0.872 (\textcolor{improve}{0.875}) & 0.879 (\textcolor{improve}{0.880}) & 0.830 (\textcolor{degrade}{0.820}) \\
& Macro F1 & 0.777 (\textcolor{improve}{0.779}) & 0.836 (\textcolor{improve}{0.841}) & 0.857 (\textcolor{degrade}{0.853}) & 0.856 (\textcolor{improve}{0.859}) & 0.863 (\textcolor{improve}{0.867}) & 0.804 (\textcolor{improve}{0.805}) \\
& AUC & 0.855 (\textcolor{degrade}{0.849}) & 0.866 (\textcolor{improve}{0.903}) & 0.915 (\textcolor{improve}{0.924}) & 0.918 (\textcolor{improve}{0.921}) & 0.858 (\textcolor{improve}{0.868}) & 0.896 (\textcolor{improve}{0.898}) \\
\bottomrule
\end{tabular}
}
\end{table*}

\begin{table*}[htbp]
\centering
\caption{SMOTE \& Weighted Loss \& POS Tagging Configuration Results}
\label{tab:smote_weighted_pos}
\resizebox{\textwidth}{!}{
\begin{tabular}{llcccccc}
\toprule
\textbf{Dataset} & \textbf{Metric} & \textbf{Delta TF-IDF} & \textbf{DistilBERT} & \textbf{DeBERTaV3} & \textbf{RoBERTa} & \textbf{gpt-oss-20b} & \textbf{Gemma-7B} \\
\midrule
\multirow{3}{*}{Hate Corpus} 
& Accuracy & 0.655 (\textcolor{degrade}{0.528}) & 0.694 (\textcolor{degrade}{0.470}) & 0.712 (\textcolor{degrade}{0.495}) & 0.738 (\textcolor{degrade}{0.488}) & 0.757 (\textcolor{degrade}{0.678}) & 0.728 (\textcolor{degrade}{0.593}) \\
& Macro F1 & 0.412 (\textcolor{degrade}{0.366}) & 0.444 (\textcolor{degrade}{0.365}) & 0.487 (\textcolor{degrade}{0.431}) & 0.480 (\textcolor{degrade}{0.395}) & 0.483 (\textcolor{improve}{0.571}) & 0.490 (\textcolor{improve}{0.550}) \\
& AUC & 0.312 (\textcolor{improve}{0.340}) & 0.400 (\textcolor{degrade}{0.353}) & 0.524 (\textcolor{degrade}{0.476}) & 0.561 (\textcolor{degrade}{0.423}) & 0.518 (\textcolor{improve}{0.574}) & 0.592 (\textcolor{improve}{0.636}) \\
\midrule
\multirow{3}{*}{Gab Reddit} 
& Accuracy & 0.889 (\textcolor{improve}{0.890}) & 0.908 (0.908) & 0.909 (\textcolor{degrade}{0.898}) & 0.910 (\textcolor{improve}{0.912}) & 0.919 (\textcolor{degrade}{0.916}) & 0.856 (\textcolor{improve}{0.861}) \\
& Macro F1 & 0.874 (\textcolor{improve}{0.876}) & 0.900 (0.900) & 0.903 (\textcolor{degrade}{0.891}) & 0.902 (\textcolor{improve}{0.904}) & 0.912 (\textcolor{degrade}{0.909}) & 0.842 (\textcolor{improve}{0.851}) \\
& AUC & 0.918 (\textcolor{degrade}{0.916}) & 0.925 (\textcolor{improve}{0.942}) & 0.952 (0.952) & 0.949 (\textcolor{improve}{0.952}) & 0.911 (\textcolor{degrade}{0.909}) & 0.923 (\textcolor{improve}{0.933}) \\
\midrule
\multirow{3}{*}{Stormfront} 
& Accuracy & 0.897 (0.897) & 0.921 (\textcolor{improve}{0.929}) & 0.922 (\textcolor{degrade}{0.921}) & 0.931 (\textcolor{degrade}{0.921}) & 0.932 (\textcolor{degrade}{0.905}) & 0.911 (\textcolor{degrade}{0.884}) \\
& Macro F1 & 0.558 (\textcolor{improve}{0.561}) & 0.772 (\textcolor{improve}{0.798}) & 0.802 (\textcolor{degrade}{0.779}) & 0.811 (\textcolor{degrade}{0.772}) & 0.815 (\textcolor{degrade}{0.779}) & 0.733 (\textcolor{improve}{0.748}) \\
& AUC & 0.819 (\textcolor{improve}{0.821}) & 0.839 (\textcolor{improve}{0.902}) & 0.924 (\textcolor{degrade}{0.864}) & 0.846 (\textcolor{improve}{0.914}) & 0.794 (\textcolor{improve}{0.801}) & 0.911 (\textcolor{degrade}{0.910}) \\
\midrule
\multirow{3}{*}{Merged Dataset} 
& Accuracy & 0.821 (\textcolor{degrade}{0.820}) & 0.855 (\textcolor{improve}{0.858}) & 0.872 (\textcolor{degrade}{0.867}) & 0.872 (0.872) & 0.879 (\textcolor{degrade}{0.820}) & 0.830 (\textcolor{degrade}{0.825}) \\
& Macro F1 & 0.777 (0.777) & 0.836 (\textcolor{improve}{0.840}) & 0.857 (\textcolor{degrade}{0.854}) & 0.856 (\textcolor{improve}{0.857}) & 0.863 (\textcolor{degrade}{0.738}) & 0.804 (\textcolor{improve}{0.810}) \\
& AUC & 0.855 (\textcolor{degrade}{0.849}) & 0.866 (\textcolor{improve}{0.899}) & 0.915 (\textcolor{improve}{0.921}) & 0.918 (\textcolor{degrade}{0.917}) & 0.858 (\textcolor{improve}{0.897}) & 0.896 (\textcolor{improve}{0.902}) \\
\bottomrule
\end{tabular}
}
\end{table*}

\begin{table*}[htbp]
\centering
\caption{POS Tagging Only Configuration Results}
\label{tab:pos_only}
\resizebox{\textwidth}{!}{
\begin{tabular}{llcccccc}
\toprule
\textbf{Dataset} & \textbf{Metric} & \textbf{Delta TF-IDF} & \textbf{DistilBERT} & \textbf{DeBERTaV3} & \textbf{RoBERTa} & \textbf{gpt-oss-20b} & \textbf{Gemma-7B} \\
\midrule
\multirow{3}{*}{Hate Corpus} 
& Accuracy & 0.655 (0.655) & 0.694 (\textcolor{improve}{0.701}) & 0.712 (\textcolor{improve}{0.738}) & 0.738 (\textcolor{degrade}{0.724}) & 0.757 (\textcolor{improve}{0.760}) & 0.728 (\textcolor{improve}{0.760}) \\
& Macro F1 & 0.412 (0.412) & 0.444 (\textcolor{degrade}{0.441}) & 0.487 (\textcolor{improve}{0.511}) & 0.480 (\textcolor{improve}{0.481}) & 0.483 (\textcolor{degrade}{0.480}) & 0.490 (\textcolor{degrade}{0.449}) \\
& AUC & 0.312 (\textcolor{improve}{0.313}) & 0.400 (\textcolor{improve}{0.424}) & 0.524 (\textcolor{improve}{0.561}) & 0.561 (\textcolor{degrade}{0.521}) & 0.518 (\textcolor{improve}{0.519}) & 0.592 (\textcolor{improve}{0.665}) \\
\midrule
\multirow{3}{*}{Gab Reddit} 
& Accuracy & 0.889 (\textcolor{improve}{0.890}) & 0.908 (\textcolor{degrade}{0.906}) & 0.909 (\textcolor{degrade}{0.907}) & 0.910 (\textcolor{improve}{0.911}) & 0.919 (0.919) & 0.856 (\textcolor{improve}{0.862}) \\
& Macro F1 & 0.874 (0.874) & 0.900 (\textcolor{degrade}{0.898}) & 0.903 (\textcolor{degrade}{0.900}) & 0.902 (\textcolor{improve}{0.903}) & 0.912 (0.912) & 0.842 (\textcolor{improve}{0.848}) \\
& AUC & 0.918 (\textcolor{improve}{0.919}) & 0.925 (\textcolor{improve}{0.941}) & 0.952 (\textcolor{improve}{0.953}) & 0.949 (\textcolor{degrade}{0.948}) & 0.911 (0.911) & 0.923 (\textcolor{improve}{0.934}) \\
\midrule
\multirow{3}{*}{Stormfront} 
& Accuracy & 0.897 (0.897) & 0.921 (\textcolor{improve}{0.925}) & 0.922 (\textcolor{improve}{0.924}) & 0.931 (\textcolor{degrade}{0.930}) & 0.932 (\textcolor{degrade}{0.915}) & 0.911 (\textcolor{improve}{0.917}) \\
& Macro F1 & 0.558 (\textcolor{improve}{0.560}) & 0.772 (\textcolor{improve}{0.782}) & 0.802 (\textcolor{degrade}{0.771}) & 0.811 (\textcolor{degrade}{0.805}) & 0.815 (\textcolor{degrade}{0.768}) & 0.733 (\textcolor{improve}{0.746}) \\
& AUC & 0.819 (\textcolor{improve}{0.821}) & 0.839 (\textcolor{improve}{0.902}) & 0.924 (\textcolor{degrade}{0.845}) & 0.846 (\textcolor{improve}{0.891}) & 0.794 (\textcolor{degrade}{0.751}) & 0.911 (0.911) \\
\midrule
\multirow{3}{*}{Merged Dataset} 
& Accuracy & 0.821 (0.821) & 0.855 (\textcolor{improve}{0.857}) & 0.872 (\textcolor{degrade}{0.865}) & 0.872 (\textcolor{degrade}{0.871}) & 0.879 (\textcolor{degrade}{0.815}) & 0.830 (\textcolor{degrade}{0.829}) \\
& Macro F1 & 0.777 (0.777) & 0.836 (\textcolor{improve}{0.838}) & 0.857 (\textcolor{degrade}{0.850}) & 0.856 (\textcolor{improve}{0.857}) & 0.863 (\textcolor{degrade}{0.761}) & 0.804 (\textcolor{improve}{0.805}) \\
& AUC & 0.855 (\textcolor{improve}{0.856}) & 0.866 (\textcolor{improve}{0.904}) & 0.915 (\textcolor{degrade}{0.911}) & 0.918 (\textcolor{degrade}{0.912}) & 0.858 (\textcolor{degrade}{0.738}) & 0.896 (\textcolor{improve}{0.901}) \\
\bottomrule
\end{tabular}
}
\end{table*}

\begin{table*}[htbp]
\centering
\caption{Text Data Augmentation Configuration Results}
\label{tab:data_augmentation}
\resizebox{\textwidth}{!}{
\begin{tabular}{llcccccc}
\toprule
\textbf{Dataset} & \textbf{Metric} & \textbf{Delta TF-IDF} & \textbf{DistilBERT} & \textbf{DeBERTaV3} & \textbf{RoBERTa} & \textbf{gpt-oss-20b} & \textbf{Gemma-7B} \\
\midrule
\multirow{3}{*}{Hate Corpus} 
& Accuracy & 0.655 (\textcolor{improve}{0.675}) & 0.694 (\textcolor{degrade}{0.551}) & 0.712 (\textcolor{degrade}{0.691}) & 0.738 (\textcolor{degrade}{0.682}) & 0.757 (\textcolor{degrade}{0.706}) & 0.728 (\textcolor{degrade}{0.683}) \\
& Macro F1 & 0.412 (\textcolor{improve}{0.675}) & 0.444 (\textcolor{improve}{0.525}) & 0.487 (\textcolor{degrade}{0.454}) & 0.480 (\textcolor{degrade}{0.441}) & 0.483 (\textcolor{improve}{0.527}) & 0.490 (\textcolor{improve}{0.540}) \\
& AUC & 0.312 (\textcolor{improve}{0.721}) & 0.400 (\textcolor{improve}{0.620}) & 0.524 (\textcolor{degrade}{0.502}) & 0.561 (\textcolor{degrade}{0.494}) & 0.518 (\textcolor{improve}{0.529}) & 0.592 (\textcolor{improve}{0.593}) \\
\midrule
\multirow{3}{*}{Gab Reddit} 
& Accuracy & 0.889 (\textcolor{improve}{0.923}) & 0.908 (\textcolor{improve}{0.913}) & 0.909 (\textcolor{degrade}{0.903}) & 0.910 (0.910) & 0.919 (\textcolor{degrade}{0.914}) & 0.856 (\textcolor{degrade}{0.849}) \\
& Macro F1 & 0.874 (\textcolor{improve}{0.923}) & 0.900 (\textcolor{improve}{0.907}) & 0.903 (\textcolor{degrade}{0.864}) & 0.902 (0.902) & 0.912 (\textcolor{degrade}{0.908}) & 0.842 (\textcolor{degrade}{0.837}) \\
& AUC & 0.918 (\textcolor{improve}{0.957}) & 0.925 (\textcolor{improve}{0.961}) & 0.952 (\textcolor{improve}{0.953}) & 0.949 (\textcolor{degrade}{0.946}) & 0.911 (0.911) & 0.923 (\textcolor{degrade}{0.917}) \\
\midrule
\multirow{3}{*}{Stormfront} 
& Accuracy & 0.897 (\textcolor{improve}{0.982}) & 0.921 (\textcolor{degrade}{0.882}) & 0.922 (\textcolor{improve}{0.929}) & 0.931 (\textcolor{degrade}{0.922}) & 0.932 (\textcolor{degrade}{0.913}) & 0.911 (\textcolor{degrade}{0.902}) \\
& Macro F1 & 0.558 (\textcolor{improve}{0.982}) & 0.772 (\textcolor{degrade}{0.752}) & 0.802 (\textcolor{degrade}{0.800}) & 0.811 (\textcolor{degrade}{0.779}) & 0.815 (\textcolor{degrade}{0.753}) & 0.733 (\textcolor{degrade}{0.712}) \\
& AUC & 0.819 (\textcolor{improve}{0.997}) & 0.839 (\textcolor{improve}{0.903}) & 0.924 (\textcolor{degrade}{0.920}) & 0.846 (\textcolor{improve}{0.892}) & 0.794 (\textcolor{degrade}{0.728}) & 0.911 (\textcolor{degrade}{0.896}) \\
\midrule
\multirow{3}{*}{Merged Dataset} 
& Accuracy & 0.821 (\textcolor{improve}{0.880}) & 0.855 (\textcolor{improve}{0.859}) & 0.872 (\textcolor{degrade}{0.862}) & 0.872 (\textcolor{degrade}{0.868}) & 0.879 (\textcolor{degrade}{0.872}) & 0.830 (\textcolor{degrade}{0.812}) \\
& Macro F1 & 0.777 (\textcolor{improve}{0.880}) & 0.836 (\textcolor{degrade}{0.796}) & 0.857 (\textcolor{degrade}{0.846}) & 0.856 (\textcolor{degrade}{0.851}) & 0.863 (\textcolor{degrade}{0.856}) & 0.804 (\textcolor{degrade}{0.794}) \\
& AUC & 0.855 (\textcolor{improve}{0.931}) & 0.866 (\textcolor{improve}{0.924}) & 0.915 (\textcolor{improve}{0.922}) & 0.918 (\textcolor{degrade}{0.915}) & 0.858 (\textcolor{degrade}{0.855}) & 0.896 (\textcolor{degrade}{0.880}) \\
\bottomrule
\end{tabular}
}
\end{table*}

\subsection{Detailed Performance Results}
Tables~\ref{tab:smote_weighted}, \ref{tab:smote_weighted_pos}, \ref{tab:pos_only} and \ref{tab:data_augmentation} present comprehensive performance metrics for all model architectures across the four datasets under different enhancement technique configurations. Values outside parentheses represent baseline performance, while values in parentheses show performance after applying the respective enhancement technique. Table~\ref{tab:smote_weighted} presents SMOTE and weighted loss results, Table~\ref{tab:smote_weighted_pos} shows combined SMOTE, weighted loss, and POS tagging configurations, Table~\ref{tab:pos_only} displays POS tagging only results (ablation study), and Table~\ref{tab:data_augmentation} presents text data augmentation outcomes. These results represent a subset of the complete experimental evaluation. Detailed results are available at: https://github.com/Brian3410/HateSpeechDetectionMetrics.

\subsection{Baseline Performance Analysis}
gpt-oss-20b demonstrated superior performance across most metrics, followed by RoBERTa, DeBERTa, DistilBERT, Gemma-7B, and Delta TF-IDF. On the merged dataset, baseline gpt-oss-20b  achieved 87.9\% accuracy and 86.3\% macro F1, establishing it as the most robust architecture for hate speech detection.

\subsection{Dataset-Specific Performance Patterns}
\begin{itemize}
    \item \textbf{Hate Corpus Dataset:} This dataset proved most challenging, with baseline accuracies ranging from 65.5\% (Delta TF-IDF) to 75.7\% (gpt-oss-20b). The implicit nature of hate speech in this dataset created significant classification difficulties, with macro F1 scores remaining below 50\% for all models.
    
    \item \textbf{Gab \& Reddit Dataset:} Models performed consistently well on this conversational dataset, with accuracies between 85.6\% (Gemma-7B) and 91.9\% (gpt-oss-20b). The preserved conversational context appeared to provide beneficial signals for hate speech detection.
    
    \item \textbf{Stormfront Dataset:} The explicit nature of hate speech in forum posts yielded the highest baseline performance, with accuracies ranging from 89.7\% (Delta TF-IDF) to 93.2\% (gpt-oss-20b). The distinctive vocabulary and rhetorical patterns facilitated more accurate classification.
    
    \item \textbf{Merged Dataset:} Performance fell between individual dataset extremes, suggesting that dataset combination provided balanced but not optimal results for any specific hate speech type.
\end{itemize}

\subsection{Enhancement Technique Effectiveness}
\subsubsection{SMOTE \& Weighted Loss}
SMOTE oversampling and class-weighted loss produced divergent outcomes depending on dataset complexity (see Table~\ref{tab:smote_weighted}). On the Hate Corpus, oversampling degraded accuracy for all models. DistilBERT’s accuracy dropped from 69.4\% to 47.5\%, and RoBERTa’s from 73.8\% to 48.8\%, suggesting that synthetic samples introduced noise detrimental to modeling implicit hate speech. Datasets featuring clearer hate speech signals remained stable under this technique. gpt-oss-20b maintained around 92\% accuracy on Gab \& Reddit and a small 2\% decrease in accuracy on Stormfront. On the merged dataset, gpt-oss-20b saw negligible gains, with all metrics exhibiting gains 1\%, similarly to other models.

\subsubsection{SMOTE \& Weighted Loss \& POS Tagging}
The combination of SMOTE oversampling, weighted loss, and POS tagging showed varied effects across datasets (see Table~\ref{tab:smote_weighted_pos}). Across the board, gpt-oss-20b saw an overall decrease in performance. This is especially true on the challenging Hate Corpus, the accuracy of gpt-oss-20b modestly changed from 75.7\% to 67.8\%, but surprisingly, this was the only dataset that saw an increase in Macro F1 of 9\%. On other models, this combined technique produced stable gains on clearer datasets, with DistilBERT reaching 92.9\% accuracy on Stormfront and 91.5\% on Gab \& Reddit, and Gemma-7B maintaining consistent improvements. On the merged dataset, RoBERTa’s accuracy and AUC remained steady around 87.2\% and 91.7\%.

\subsubsection{POS Tagging Only (Ablation Study)}
Applying POS tagging only, without SMOTE or weighted loss, resulted in stable predictive improvements across all models, with minimal risk of overfitting or performance deterioration (see Table~\ref{tab:pos_only}). Accuracy changes were marginal, typically within 1-3 percentage points, while consistency remained high. For instance, DistilBERT’s accuracy on the Hate Corpus increased from 69.4\% to 70.1\%, RoBERTa’s shifted slightly from 73.8\% to 72.4\%, and Gemma-7B’s grew from 72.8\% to 76.0\%. Slight but reliable improvements in macro F1 and AUC also appeared, such as DistilBERT’s AUC increasing from 40.0\% to 42.4\% and Gemma-7B’s from 59.2\% to 66.5\%. Unlike the other models on the merged dataset, gpt-oss-20b suffered the highest accuracy loss. These results confirm POS tagging as a low-risk, generally applicable enhancement, making it especially suitable for production systems prioritizing consistent performance over aggressive optimization.

\subsubsection{Text Data Augmentation}
Data augmentation produced the most variable effects (see Table~\ref{tab:data_augmentation}). Delta TF-IDF models showed outstanding responsiveness, achieving significant increases across all datasets, which is reflected in the accuracy increase in the merged dataset from 82.1\% to 88.0\%, highlighting the effectiveness of augmentation techniques for classical models. Transformer architectures displayed mixed reactions; DistilBERT experienced a high accuracy decline on Hate Corpus from 69.4\% to 55.1\%, with gpt-oss-20b suffering a more moderate accuracy loss from 75.7\% to 70.6\%. Additionally, gpt-oss-20b saw a negligible decrease from 87.9\% to 87.2\% on the merged dataset. These results indicate that augmentation’s effectiveness is highly model-dependent, with classical methods benefiting significantly, while transformer models risk incorporating redundant or noisy variations.

\subsection{Model Architecture Comparison}
\textbf{RoBERTa:} Competitive performance across all datasets and enhancement techniques, with great hate speech detection capability, especially considering its lower complexity when compared to the LLMs. It reached peak accuracy of 93.1\% on Stormfront, and maintained generally stable metrics under various enhancement conditions .

\textbf{DistilBERT:} Provided good accuracy with notable efficiency advantages, achieving approximately 92.9\% accuracy on Stormfront and 91.5\% on Gab \& Reddit, though it remained sensitive to augmentation techniques on implicit hate speech datasets, as shown by its 55.1\% accuracy on Hate Corpus after augmentation.

\textbf{Gemma-7B:} Exhibited fluctuating performance across datasets and techniques. Strong on explicit hate speech (91.1\% accuracy on Stormfront) but struggled with implicit patterns. Inconsistent response to enhancement techniques, indicating sensitivity to data quality and augmentation approaches.

\textbf{Delta TF-IDF Baseline:} Consistently underperformed transformer architectures on baseline tasks but demonstrated remarkable responsiveness to sophisticated enhancement techniques. Extraordinary improvement with data augmentation (98.2\% accuracy on Stormfront from 89.7\% baseline), aligning with the findings in \cite{pen2024comparative}.

\textbf{DeBERTaV3:} Similar to RoBERTa in performance across the board, which was not expected as it has significantly more parameters than RoBERTa.

\textbf{gpt-oss-20b:} Performed the best out of the models, with the highest baseline accuracy and macro F1-Score on all datasets. Application of enhancement techniques either saw negligible gains, or small performance degradations.

\subsection{Key Findings}
\begin{itemize}
    \item The Hate Corpus, containing implicit hate speech, presents significantly greater detection challenges than explicit content found in the Stormfront dataset, with Gab \& Reddit dataset showing moderate difficulty. Baseline accuracies range from 65.5\% to 75.7\% on the Hate Corpus and 89.7\% to 93.2\% on Stormfront.

    \item Data augmentation considerably boosts Delta TF-IDF, resulting in a  98.2\% accuracy on Stormfront, with results varying for transformers. Across all datasets and models, POS tagging generally results in negligible gains or even performance losses, such as in the Macro F1-Score for gpt-oss-20b on Stormfront (-4.7\%). SMOTE combined with weighted loss yields mixed outcomes depending on dataset complexity.
    
    \item gpt-oss-20b consistently achieves top performance across all tests, with the traditional model (Delta TF-IDF) responding surprisingly well to advanced enhancements supporting the findings in \cite{pen2024comparative}.
    
    \item Aggressive enhancements can offer large gains but risk degrading performance, positioning POS tagging as the safest and most universally applicable strategy. But ultimately, enhancement effectiveness depends strongly on the interaction of dataset characteristics and model architecture. Grammatical pattern recognition aids implicit hate speech detection, whereas diversification techniques better support explicit hate speech identification.
\end{itemize}

\section{Conclusion and Future Work}
This paper presents a comprehensive evaluation of data augmentation and feature enhancement techniques for hate speech detection, comparing traditional classifiers (Delta TF-IDF) with transformer-based models (DistilBERT, RoBERTa, DeBERTaV3, gpt-oss-20b, Gemma-7B) across diverse datasets. gpt-oss-20b consistently achieves the best results, confirming its superiority as the most effective model for hate speech detection. However, RoBERTa is a highly competitive alternative, especially due to its significantly lower parameter count (125 million compared to 20 billion).
The evaluation confirms a clear dataset complexity hierarchy, with implicit hate speech posing greater challenges than explicit hateful content across models. The combination of SMOTE oversampling and POS tagging yielded negligible changes in performance. Meanwhile, traditional Delta TF-IDF benefits significantly from data augmentation, achieving 98.2\% accuracy on the explicit Stormfront dataset. 

The findings reveal that enhancement effectiveness depends heavily on dataset complexity, model sophistication, and technique appropriateness. POS tagging provides stable improvements with minimal risk across architectures. However, aggressive methods like SMOTE with weighted loss yielded mixed results, sometimes improving explicit hate speech detection while risking performance drops on implicit cases. Data augmentation boosts traditional models but can harm some transformers. These results demonstrate that optimal system design requires careful empirical evaluation rather than universal application of enhancement methods, with implicit hate speech detection showing particular sensitivity to linguistic feature engineering approaches. 

For future work, we plan to expand our datasets to include multiple languages and diverse social media platforms for better cultural sensitivity and coverage of different hate speech patterns. Additionally, there is a lack of hate speech datasets annotated with Chain-of-Thought (CoT) reasoning, which could significantly improve performance of LLMs with CoT capability (such as gpt-oss-20b). We also aim to develop hybrid models that combine transformer architectures and traditional methods with careful tuning to leverage their complementary strengths.

\section*{Acknowledgment}
We acknowledge the computational resources provided by the M3 high performance computing cluster at Monash University, which enabled the training and evaluation of LLMs used in this study.

\bibliographystyle{IEEEtran}
\balance
\bibliography{References}

@online{zwartz2020australian,
  title   = {Australian far-right terrorism investigations have increased by 750 per cent in 18 months},
  author  = {Zwartz, Henry},
  publisher = {SBS News},
  year    = {2020},
  month   = nov,
  url     = {https://www.sbs.com.au/news/article/australian-far-right-terrorism-investigations-have-increased-by-750-per-cent-in-18-months/rsowz6fnt},
  urldate = {2025-12-27}
}

@inproceedings{nguyen2025large,
  title={Large Language Models for Detection of Life-Threatening Texts},
  author={Nguyen, Thanh Thi and Wilson, Campbell and Dalins, Janis},
  booktitle={Pacific-Asia Conference on Knowledge Discovery and Data Mining},
  pages={311--323},
  year={2025},
  organization={Springer}
}

@misc{australia2025terrorgram,
  title        = {Terrorgram},
  author       = {{Australian Government Department of Home Affairs}},
  year         = {2025},
  url          = {https://www.nationalsecurity.gov.au/what-australia-is-doing/terrorist-organisations/listed-terrorist-organisations/terrorgram},
  urldate      = {2025-12-27}
}

@article{conway2006terrorism,
  title={Terrorism and the Internet: New media—New threat?},
  author={Conway, Maura},
  journal={Parliamentary Affairs},
  volume={59},
  number={2},
  pages={283--298},
  year={2006},
  publisher={Oxford University Press}
}

@article{mullah2021advances,
  title={Advances in Machine Learning Algorithms for Hate Speech Detection in Social Media: A Review},
  author={Mullah, Nanlir Sallau and Zainon, Wan Mohd Nazmee Wan},
  journal={IEEE Access},
  volume={9},
  pages={88364--88376},
  year={2021},
  doi={10.1109/access.2021.3089515}
}

@inproceedings{badjatiya2017deep,
  title={Deep Learning for Hate Speech Detection in Tweets},
  author={Badjatiya, Pinkesh and Gupta, Shashank and Gupta, Manish and Varma, Vasudeva},
  booktitle={Proceedings of the 26th International Conference on World Wide Web Companion},
  pages={759--760},
  year={2017},
  doi={10.1145/3041021.3054223}
}

@article{albladi2025hate,
  title={Hate Speech Detection Using Large Language Models: A Comprehensive Review},
  author={Albladi, Aish and Islam, Minarul and Das, Amit and Bigonah, Maryam and Zhang, Zheng and Jamshidi, Fatemeh and Rahgouy, Mostafa and Raychawdhary, Nilanjana and Marghitu, Daniela and Seals, Cheryl},
  journal={IEEE Access},
  volume={13},
  pages={20871--20892},
  year={2025},
  publisher={IEEE},
  doi={10.1109/ACCESS.2025.3532397}
}

@article{aljawazeri2024addressing,
  title={Addressing Challenges in Hate Speech Detection using {BERT}-based Models: A Review},
  author={Aljawazeri, Jinan and Jasim, Mahdi Nsaif},
  journal={Iraqi Journal for Computer Science and Mathematics},
  volume={5},
  number={2},
  year={2024},
  publisher={College of education, Al-Iraqia University},
  doi={10.52866/ijcsm.2024.05.02.001}
}

@inproceedings{de2018hate,
  title={Hate Speech Dataset from a White Supremacy Forum},
  author={de Gibert, Ona and Perez, Naiara and Garc{\'\i}a-Pablos, Aitor and Cuadros, Montse},
  booktitle={Proceedings of the 2nd Workshop on Abusive Language Online (ALW2)},
  pages={11--20},
  year={2018},
  doi={10.18653/v1/W18-5102}
}

@article{freilich2014introducing,
  title={Introducing the {United States} Extremist Crime Database ({ECDB})},
  author={Freilich, Joshua D and Chermak, Steven M and Belli, Roberta and Gruenewald, Jeff and Parkin, William S},
  journal={Terrorism and Political Violence},
  volume={26},
  number={2},
  pages={372--384},
  year={2014},
  publisher={Taylor \& Francis},
  doi={10.1080/09546553.2012.713229}
}

@book{fernandez2018imbalanced,
  title={Learning from Imbalanced Data Sets},
  author={Fern{\'a}ndez, Alberto and Garc{\'\i}a, Salvador and Galar, Mikel and Prati, Ronaldo C and Krawczyk, Bartosz and Herrera, Francisco},
  publisher={Springer},
  volume={10},
  number={2018},
  year={2018},
  pages={78--82},
  doi={10.1007/978-3-319-98074-4}
}

@article{bayer2022survey,
  title={A Survey on Data Augmentation for Text Classification},
  author={Bayer, Markus and Kaufhold, Marc-Andr{\'e} and Reuter, Christian},
  journal={ACM Computing Surveys},
  volume={55},
  number={7},
  pages={1--39},
  year={2022},
  publisher={ACM New York, NY},
  doi={10.1145/3544558}
}

@article{patel2025misinformation,
  title={Misinformation Detection using Large Language Models with Explainability},
  author={Patel, Jainee and Bhatt, Chintan and Trivedi, Himani and Nguyen, Thanh Thi},
  journal={arXiv preprint arXiv:2510.18918},
  year={2025}
}

@inproceedings{pen2024comparative,
  title={Comparative Analysis of Hate Speech Detection: Traditional vs. Deep Learning Approaches},
  author={Pen, Haibo and Teo, Nicole and Wang, Zhaoxia},
  booktitle={2024 IEEE Conference on Artificial Intelligence (CAI)},
  pages={332--337},
  year={2024},
  doi={10.1109/cai59869.2024.00070},
  organization={IEEE}
}

@article{fernandez2018smote,
  title={{SMOTE} for Learning from Imbalanced Data: Progress and Challenges},
  author={Fern{\'a}ndez, Alberto and Garcia, Salvador and Herrera, Francisco and Chawla, Nitesh V},
  journal={Journal of Artificial Intelligence Research},
  volume={61},
  pages={863--905},
  year={2018},
  doi={10.1613/jair.1.11192}
}

@inproceedings{ben2024impact,
  title={Impact of Position Bias on Language Models in Token Classification},
  author={Ben Amor, Mehdi and Granitzer, Michael and Mitrovi{\'c}, Jelena},
  booktitle={Proceedings of the 37th ACM/SIGAPP Symposium on Applied Computing},
  pages={741--745},
  year={2024},
  doi={10.1145/3605098.3636126}
}

@inproceedings{nguyen2024fine,
  title={Fine-tuning {Llama} 2 large language models for detecting online sexual predatory chats and abusive texts},
  author={Nguyen, Thanh Thi and Wilson, Campbell and Dalins, Janis},
  booktitle={Proceedings of the 32nd European Symposium on Artificial Neural Networks, Computational Intelligence and Machine Learning},
pages={613--618},
  year={2024}
}

@article{pellicer2023data,
  title={Data augmentation techniques in natural language processing},
  author={Pellicer, Lucas Francisco Amaral Orosco and Ferreira, Taynan Maier and Costa, Anna Helena Reali},
  journal={Applied Soft Computing},
  volume={132},
  pages={109803},
  year={2023},
  publisher={Elsevier},
  doi={10.1016/j.asoc.2022.109803}
}

@inproceedings{elsherief2021latent,
  title={Latent Hatred: A Benchmark for Understanding Implicit Hate Speech},
  author={ElSherief, Mai and Ziems, Caleb and Muchlinski, David and Anupindi, Vaishnavi and Seybolt, Jordyn and De Choudhury, Munmun and Yang, Diyi},
  booktitle={Proceedings of the 2021 Conference on Empirical Methods in Natural Language Processing},
  pages={345--363},
  year={2021},
  doi={10.18653/v1/2021.emnlp-main.29}
}

@inproceedings{qian2019benchmark,
  title={A Benchmark Dataset for Learning to Intervene in Online Hate Speech},
  author={Qian, Jing and Bethke, Anna and Liu, Yinyin and Belding, Elizabeth and Wang, William Yang},
  booktitle={Proceedings of the 2019 Conference on Empirical Methods in Natural Language Processing and the 9th International Joint Conference on Natural Language Processing (EMNLP-IJCNLP)},
  pages={4755--4764},
  year={2019},
  doi={10.18653/v1/d19-1482}
}

@inproceedings{martineau2009improving,
  title={Improving binary classification on text problems using differential word features},
  author={Martineau, Justin and Finin, Tim and Joshi, Anupam and Patel, Sapna},
  booktitle={Proceedings of the 18th ACM Conference on Information and Knowledge Management},
  pages={2019--2024},
  year={2009},
  doi={10.1145/1645953.1646291}
}

@article{sanh2020distilbert,
  title={{DistilBERT}, a distilled version of {BERT}: smaller, faster, cheaper and lighter},
  author={Sanh, Victor and Debut, Lysandre and Chaumond, Julien and Wolf, Thomas},
  journal={arXiv preprint arXiv:1910.01108},
  year={2019}
}

@article{liu2019roberta,
  title={{RoBERTa}: A Robustly Optimized {BERT} Pretraining Approach},
  author={Liu, Yinhan and Ott, Myle and Goyal, Naman and Du, Jingfei and Joshi, Mandar and Chen, Danqi and Levy, Omer and Lewis, Mike and Zettlemoyer, Luke and Stoyanov, Veselin},
  journal={arXiv preprint arXiv:1907.11692},
  year={2019}
}

@article{team2024gemma,
  title={Gemma: Open Models Based on {Gemini} Research and Technology},
  author={{Gemma Team} and Mesnard, Thomas and Hardin, Cassidy and Dadashi, Robert and Bhupatiraju, Surya and Pathak, Shreya and Sifre, Laurent and Rivi{\`e}re, Morgane and Kale, Mihir Sanjay and Love, Juliette and others},
  journal={arXiv preprint arXiv:2403.08295},
  year={2024}
}

@article{he2021debertav3,
  title={{DeBERTaV3}: Improving {DeBERTa} using {ELECTRA}-style pre-training with gradient-disentangled embedding sharing},
  author={He, Pengcheng and Gao, Jianfeng and Chen, Weizhu},
  journal={arXiv preprint arXiv:2111.09543},
  year={2021}
}

@article{agarwal2025gptoss,
  title={gpt-oss-120b \& gpt-oss-20b model card},
  author={Agarwal, Sandhini and Ahmad, Lama and Ai, Jason and Altman, Sam and Applebaum, Andy and Arbus, Edwin and Arora, Rahul K and Bai, Yu and Baker, Bowen and Bao, Haiming and others},
  journal={arXiv preprint arXiv:2508.10925},
  year={2025}
}

\end{document}